\documentclass{article}
\usepackage{spconf,amsmath,graphicx}
\usepackage{multirow}
\usepackage{amsfonts}
\usepackage{gensymb}
\usepackage{enumitem}
\setlist{nosep, leftmargin=14pt}

\usepackage{mwe} 

\title{Contouring by Unit Vector Field Regression}
\name{Amir Jamaludin$^{\star}$ \qquad Sarim Ather$^{\dagger}$ \qquad Timor Kadir$^{\ddagger}$ \qquad Rhydian Windsor$^{\star}$ }

\address{$^{\star}$ Visual Geometry Group, Department of Engineering Science, University of Oxford \\  $^{\dagger}$ Oxford University Hospitals NHS Foundation Trust \\  $^{\ddagger}$ Plexalis Ltd}

\begin{document}
%
\maketitle
\begin{abstract}
This work introduces a simple deep-learning based method to delineate contours by `walking' along learnt unit vector fields. We demonstrate the effectiveness of our pipeline on the unique case of open contours on the task of delineating the sacroiliac joints (SIJs) in spinal MRIs. We show that: (i) $95\%$ of the time the average root mean square error of the predicted contour against the original ground truth is below 4.5 pixels (2.5mm for a standard T1-weighted SIJ MRI), and (ii) the proposed method is better than the baseline of regressing vertices or landmarks of contours.
\end{abstract}

\vspace{-1.0em}

\section{Introduction}
\label{sec:intro}

Contouring objects is a very important step in various medical image analysis tasks. Currently, one common approach is to predict a segmentation map of the object and then extract the map's edges. However, this approach has limitations. Firstly, the output segmentations are not necessarily a single interconnected volume and thus additional post-processing is required before finding edges, which can introduce errors (e.g. by removing additional volumes). Secondly, this method does not allow for detecting open contours. An alternative approach is to treat pixels along the open contour as segmentation targets. However, this approach often leads to small, challenging segmentation targets. Furthermore, these approaches do not guarantee a unique solution or easily allow for sub-pixel precision contours in both the open and closed settings.

Therefore, in this paper, we propose a new method to delineate contours, avoiding these limitations. This is done by `walking' along a learnt vector field. Along the contour, the field should point parallel to the contour, whereas outside the contour the field should point to the nearest contour point. To demonstrate the effectiveness of this method, we apply it to a novel task; delineating the sacroiliac joint (SIJ) boundary in clinical  MRI scans. 

\hfill \break
\noindent \textbf{Sacroiliac Joint Delineation.} The SIJ is the joint between the sacrum of the spine and the ilium bones of the pelvis. There are two SIJs per person, one on the left and one on the right. MR imaging is typically done to look at the inflammation of the SIJ, or sacroiliitis, which is one of the causes of low back pain and part of the diagnosis for ankylosing spondylitis (AS). In AS, the severity of SIJ inflammation is used to assess disease progression. AS Grading systems often refer to specific regions surrounding the SIJ  \cite{Krohn14}, which makes SIJ detection a must. Since the SIJ is defined as the space between two bones, we follow the approach suggested by \cite{maksymowych2009} and delineate each SIJ as an individual open contour, which is beneficial for the further downstream task of grading the SIJ.

\begin{figure}[!t]
  \centering
  \centerline{\includegraphics[width=\linewidth]{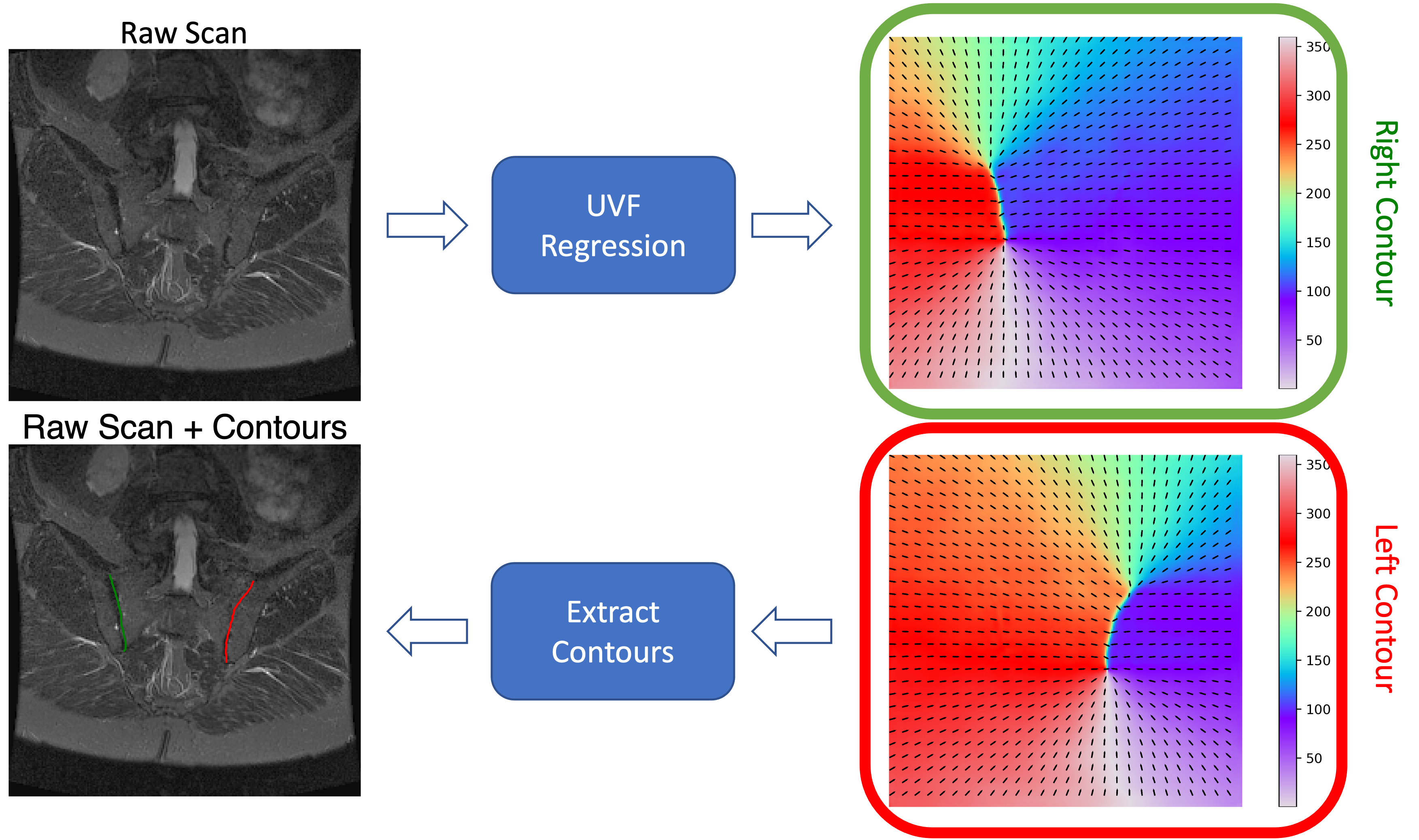}}
\caption{Overview of the contouring pipeline on an example SIJ MRI. The model outputs two vector fields, one for both the left (red) and right (green) SIJs. Each vector field is shown as a gradient map of the angle (in degrees) of the vector at that point. These vector fields are then used to extract contours for both the SIJs, shown in the bottom left panel.}
\vspace{-1.5em}
\label{fig:overview}
\end{figure}

\hfill \break
\noindent \textbf{Related Work.} There have been multiple works on detecting or segmenting parts of the spine in spinal medical imaging across several imaging modalities, e.g. intervertebral discs~\cite{Jamaludin15} and vertebral bodies in MRI~\cite{Windsor20} and CT scans~\cite{tao22} as well as the whole spine in DXA scans~\cite{jamaludin18, Bourigault20}. However, there has been relatively little research on detecting the SIJ and related downstream tasks, for example, inflammation prediction or quantifying structural changes. The closest work to date on SIJ delineation is \cite{Faleiros20}. However, this method focuses on the classification of sacroiliitis and requires manual annotation to locate the SIJ region. Another closely related work is \cite{Bressem22}, where the authors propose a method to detect changes in the SIJ. However, this is done without explicitly focusing on the SIJ region, instead taking the whole slice of an SIJ MRI as input. We propose that by delineating the SIJ, models can focus on the exact region of the disease without additional noise from surrounding anatomical structures.  

Our contouring method has analogies to several works on shape representation using deep learning via implicit functions (e.g.~\cite{Mescheder19,Park19,Chibane20}). In this case, rather than representing shapes as a binary mask over a regular grid of voxels, a model learns $f : \mathbb{R}^3 \rightarrow \mathbb{R}$,
such that $f(x,y,z)$ estimates the closest distance from point $(x,y,z)$ to the object of interest's surface (\textit{signed distance} functions), or whether $(x,y,z)$ is occupied by the shape (\textit{occupancy} functions). These methods allow for sub-pixel/voxel precision representations of surfaces. Though we validated our approach on SIJ MRIs, it is worth noting that open contours are widely used in other medical imaging tasks e.g. torso contour segmentation for better ECG interpretation \cite{smith2022}, and reconstructing 3D meshed of the heart from 2D cardiac MRIs \cite{banerjee2021}.


\vspace{-0.25em}
\section{Approach Overview}
\label{sec:method}
Our method takes as input 2D images and outputs an array 
of vertices delineating the contour of interest. This is done by a two stage approach: (i) Firstly, a model to predicts \textit{a unit vector field} (UVF) for the image.  At location $\mathbf{x}$, the UVF indicates the direction towards the nearest point on the contour  of interest (ii) Secondly, we propose a method to extract open contours from this learned vector field. Our overall approach  for the task of SIJ delineation 
can be seen in Figure~\ref{fig:overview}.

\begin{figure}[!t]
\begin{minipage}[b]{.48\linewidth}
  \centering
  \centerline{\includegraphics[height=3.6cm]{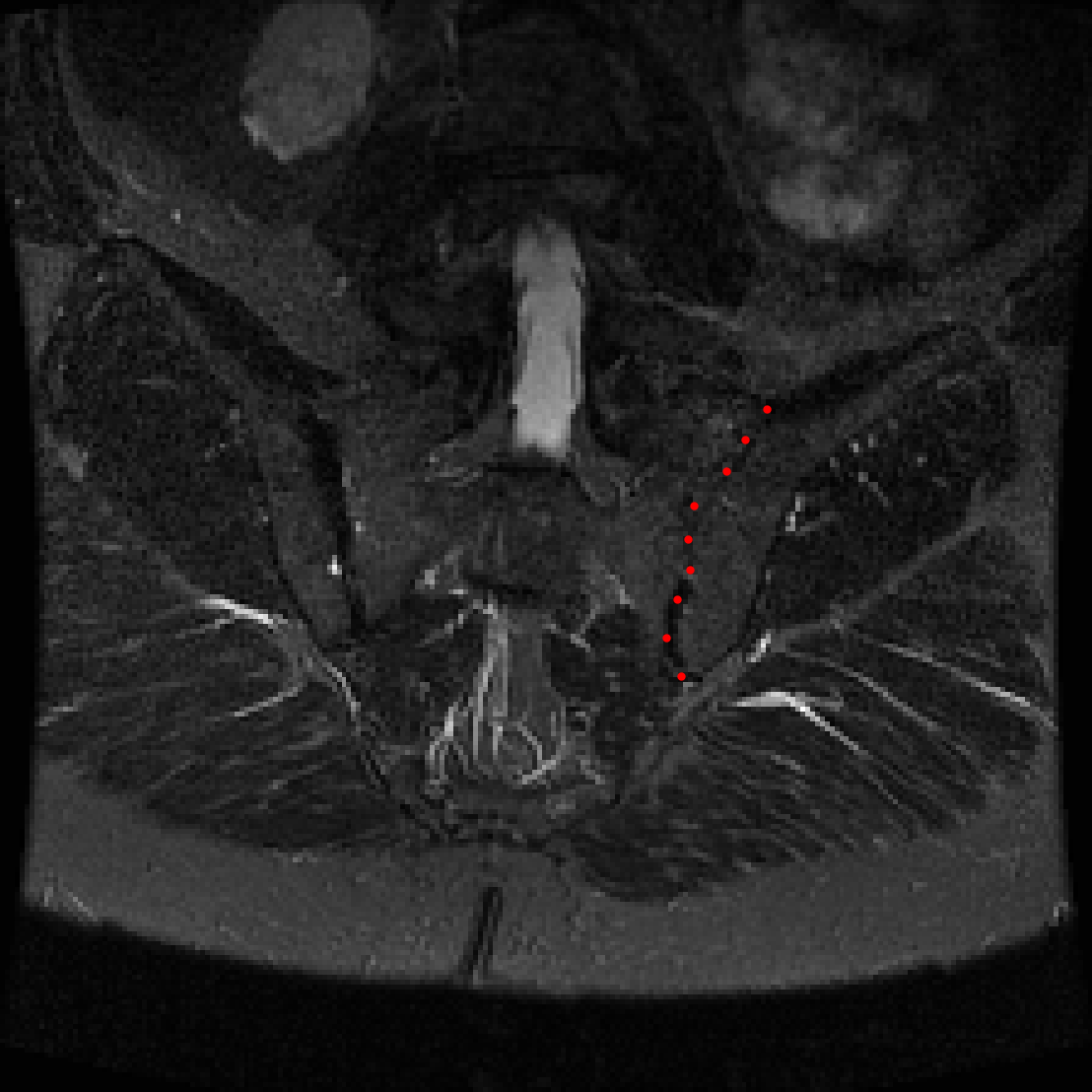}}
  \centerline{(a) Target Vertices}\medskip
\end{minipage}
 \hspace{0.2cm}
\begin{minipage}[b]{0.48\linewidth}
  \centering
  \centerline{\includegraphics[height=3.6cm]{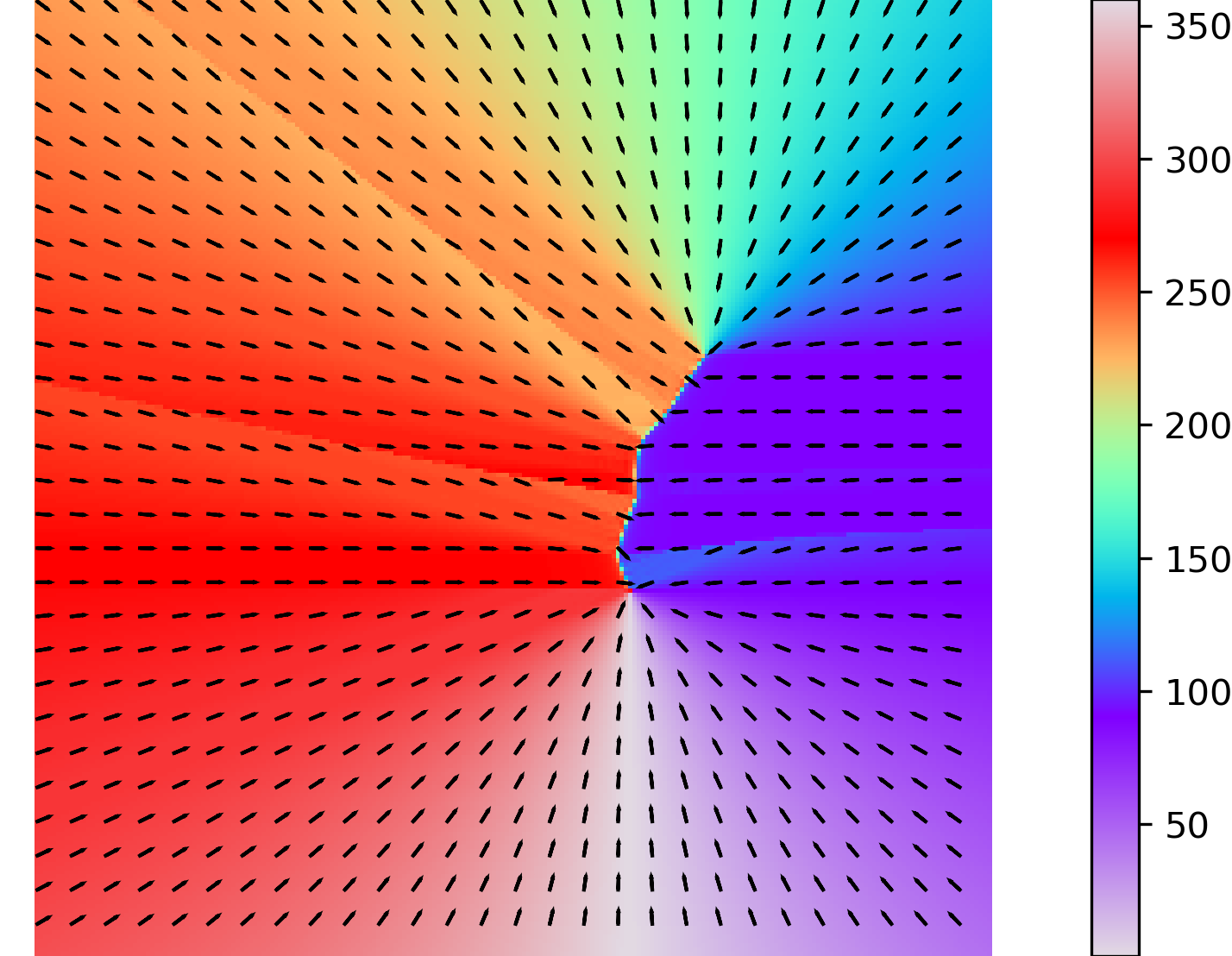}}
  \centerline{(b) Target UVF}\medskip
\end{minipage}

\caption{The Unit Vector Field (UVF): (a) a slice of an SIJ MRI with annotated landmarks in red delineating the left SIJ (with respect to the patient), (b) the resulting target UVF, overlaid on top of a gradient map of the field's direction in degrees. 
}
\vspace{-1.0em}
\label{fig:uvf}
\end{figure}

\subsection{Unit Vector Fields}
\label{ssec:subhead}
The idea of contours and vector fields in combination is not a new one; for example, several early works in computer vision combined Snakes \cite{kass88} with gradient vector flow \cite{xu98}, i.e. a vector field pointing towards object edges in a given image. However, instead of defining the vector field using object edges, we instead learn the unit vector field, $\mathbf{\hat{v}}_{i,j}$, where at each location in the vector field, $(i,j)$, the field `points' to the nearest vertex, i.e.\ annotated ground truth landmark, on the contour of the object. The unit vector field is made of two separate  x and y components corresponding to the directions of the vectors in the field. To preserve the directionality of the contour, we impose a rule where vectors laying on top of the contour should `point' to where the next vertex is expected. An example unit vector field can be seen in Figure~\ref{fig:uvf}.

\begin{figure}[!t]
\begin{minipage}[b]{.48\linewidth}
  \centering
  \centerline{\includegraphics[height=3.6cm]{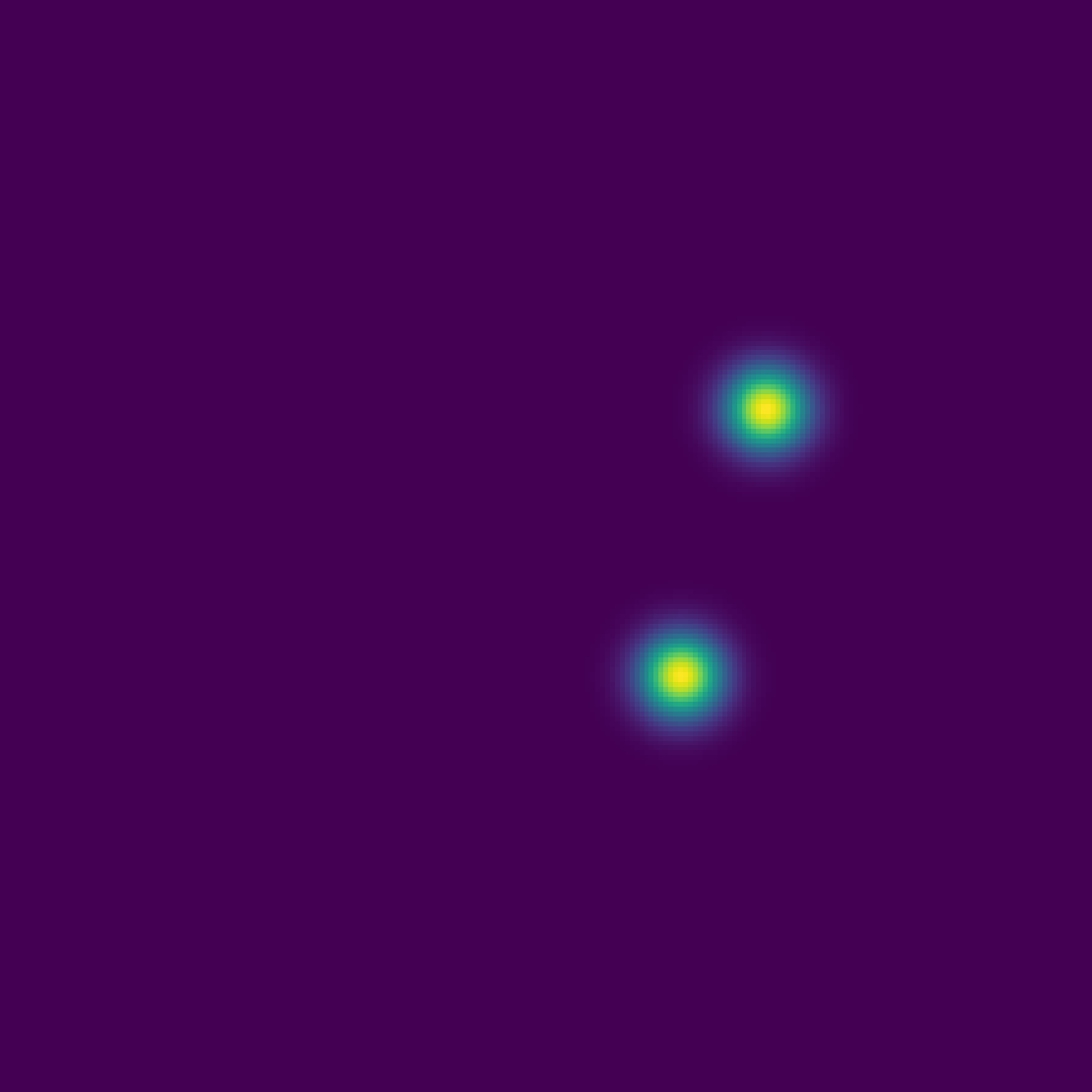}}
  \centerline{(a) Gaussian Heatmaps}\medskip
\end{minipage}
\begin{minipage}[b]{0.48\linewidth}
  \centering
  \centerline{\includegraphics[height=3.6cm]{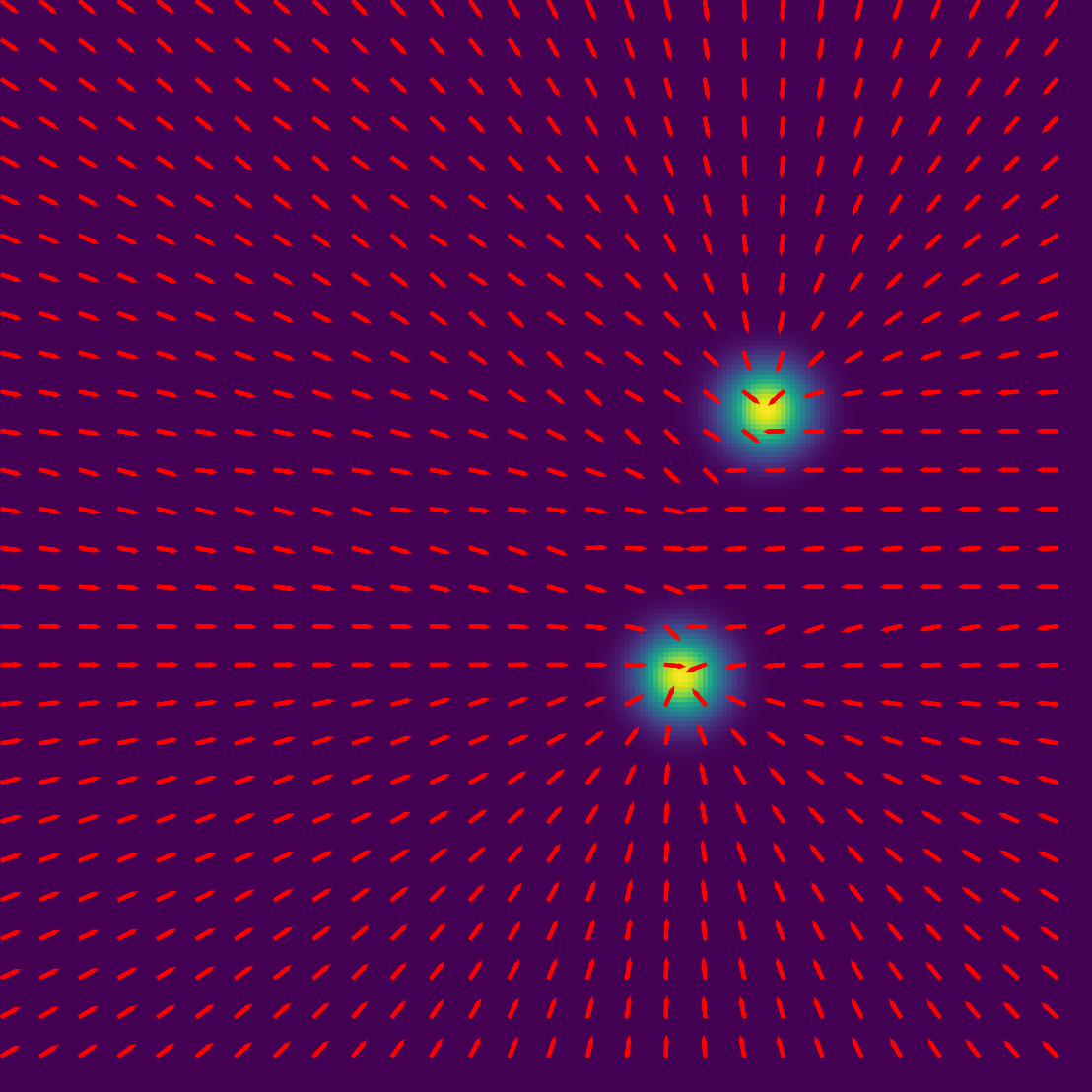}}
  \centerline{(b) Gaussians + UVF }\medskip
\end{minipage}
\begin{minipage}[b]{0.48\linewidth}
  \centering
  \centerline{\includegraphics[height=3.6cm]{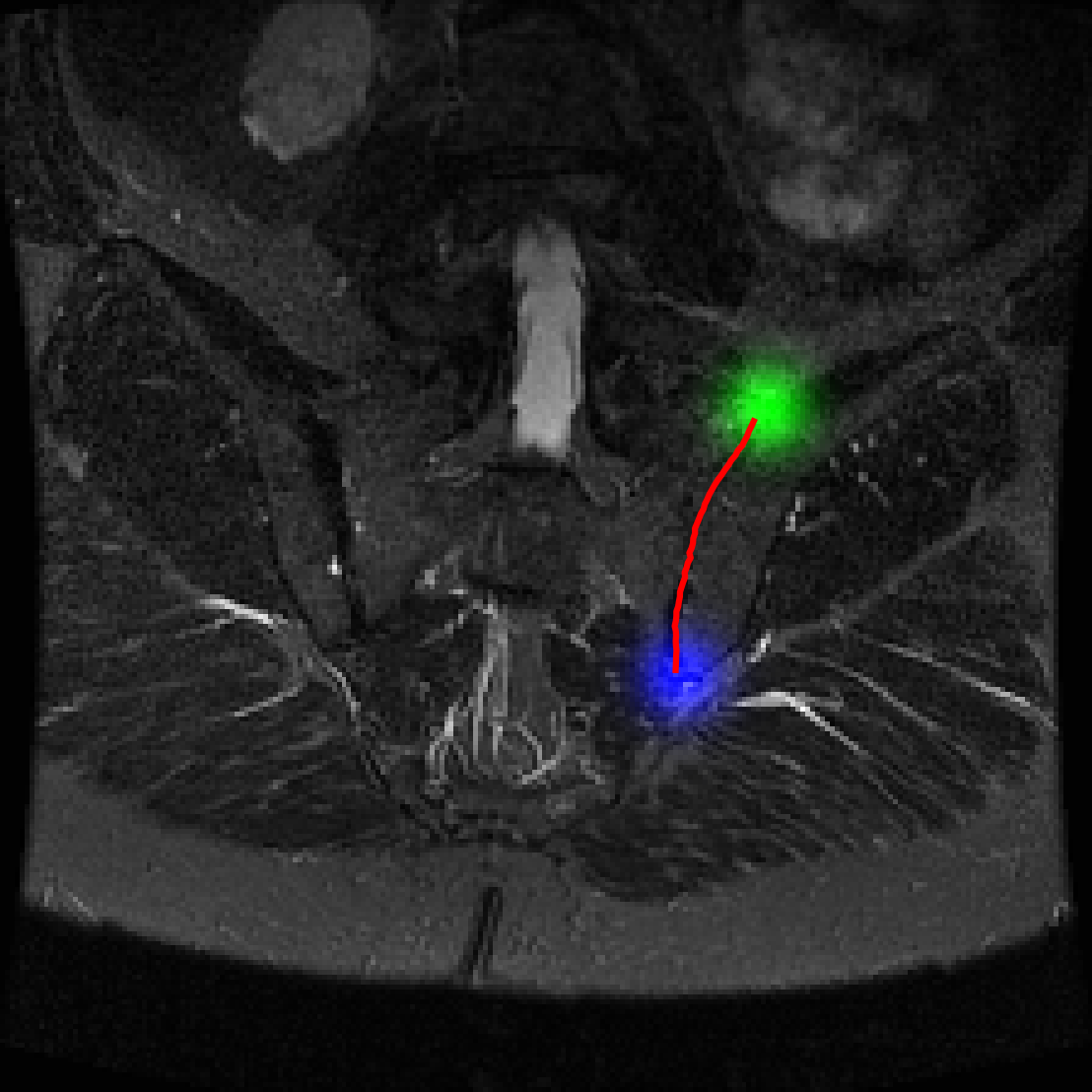}}
  \centerline{(c) Gaussian + Contour}\medskip
\end{minipage}
 \hspace{0.2cm}
\begin{minipage}[b]{0.48\linewidth}
  \centering
  \centerline{\includegraphics[height=3.6cm]{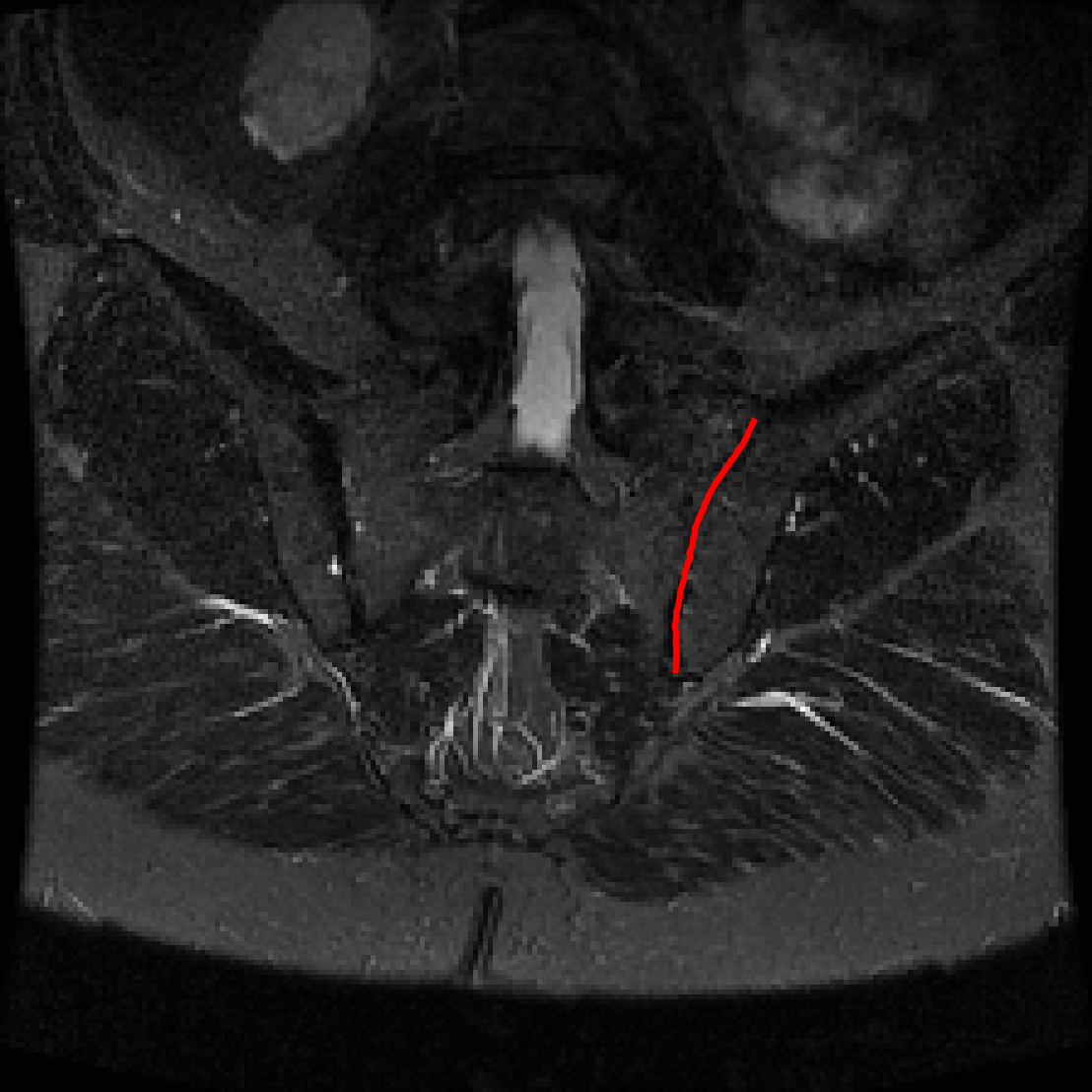}}
  \centerline{(d) Original Image + Contour}\medskip
\end{minipage}

\caption{Following on from the example shown in Figure~\ref{fig:uvf}, alongside the UVF, we regress two 2D Gaussian heatmaps. (a) 2 Gaussians representing the start and end points of the contour, (b) the UVF overlaid on top of the Gaussians, (c) the contour which starts from the Gaussian now marked in Green and ends on the Gaussian marked in Blue, (d) the final contour for the left SIJ marked in Red.}
\vspace{-1.0em}
\label{fig:ctr}
\end{figure}

\subsection{Extracting Contours From Unit Vector Fields}
\label{ssec:startend}
The unit vector field alone does not obviously indicate where a contour starts and ends. We solve this by also predicting the start and end points with the same network that generates the unit vector field; this is done simultaneously as a separate output. We take inspiration from previous works \cite{Windsor20,Payer15,Zhou2019} and regress two distinct Gaussian heatmaps for the start and end points respectively. Each Gaussian has a maximum value of 1 and a variance proportional to the area of the task-relevant object. In our case, we use the sacrum, i.e.  the area which lies in between two SIJs. In the case where the contour is without a defined area of interest, we suggest scaling the Gaussian heatmap proportional to the length of the overall contour. The beginning of the contour is defined from the Gaussian heatmap designated as the start point. We then iteratively `walk' following the direction in the UVF, $\mathbf{\hat{v}}_{i,j}$, and the contour ends when approaching the second Gaussian heatmap i.e. end point. Each step is 1 unit in magnitude, although this could be adjusted to generate contours of varying fidelity.
Figure~\ref{fig:ctr} gives an example of how a contour is defined with the Gaussian heatmaps and the UVF. 
Since the UVF can be visualized, errors can be more easily interpreted. 
Though not shown in this work, a closed contour solution would not require heatmaps and could be found by simply searching for a loop in the UVF.

\begin{figure}[!t]
\begin{minipage}[b]{.32\linewidth}
  \centering
  \centerline{\includegraphics[height=2.4cm]{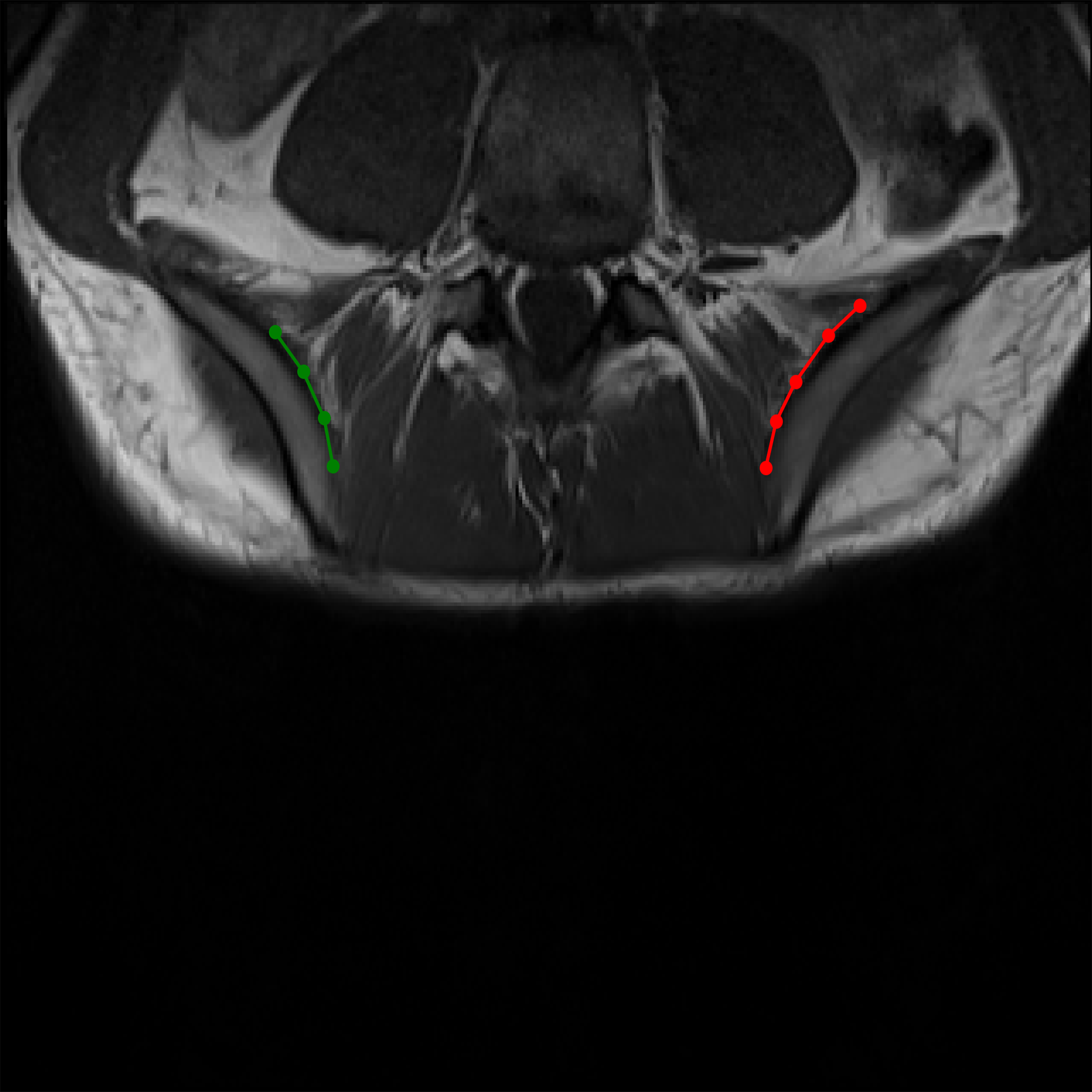}}
  \centerline{(a) Ant. T1-w}\medskip
\end{minipage}
\begin{minipage}[b]{0.32\linewidth}
  \centering
  \centerline{\includegraphics[height=2.4cm]{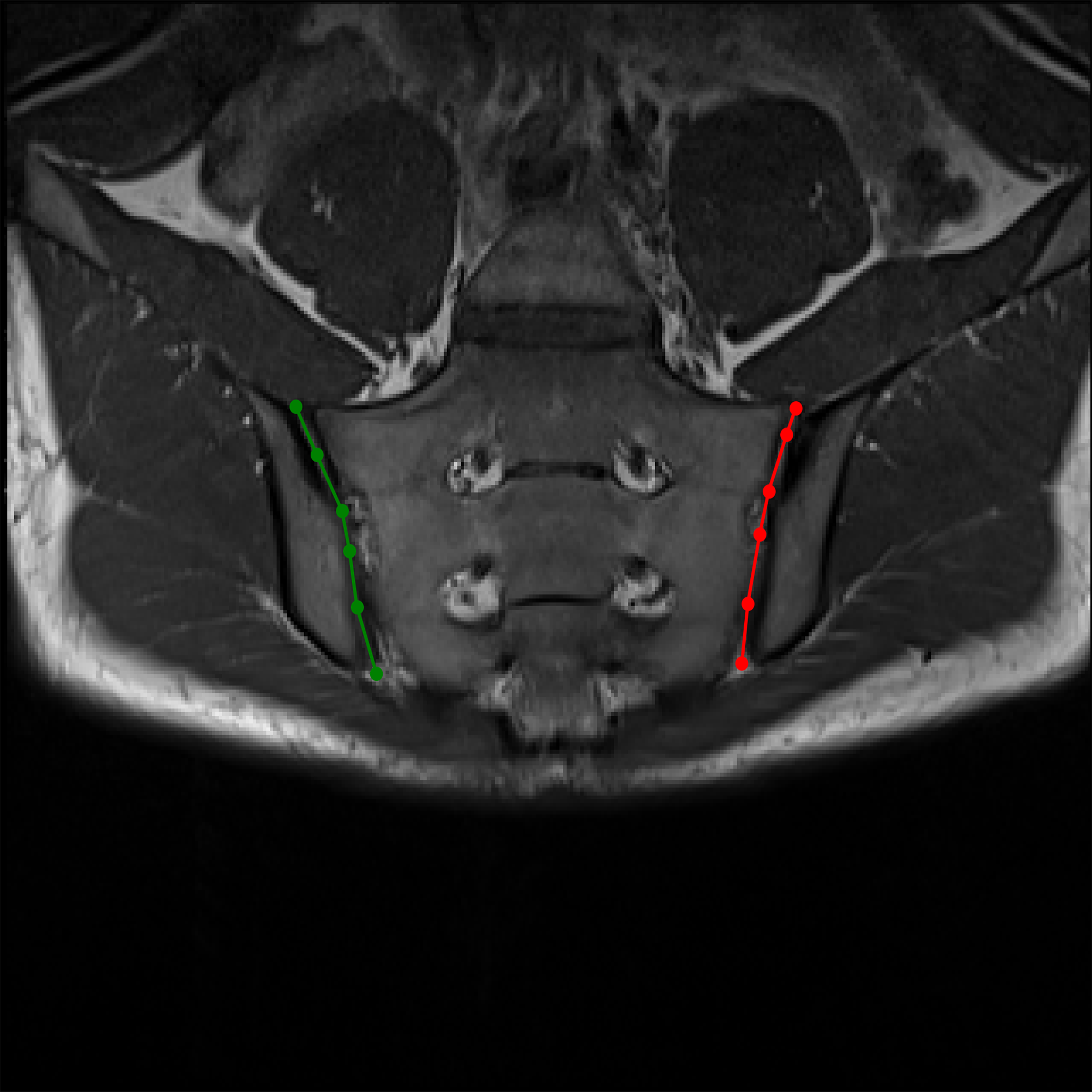}}
  \centerline{(b) Mid-cor. T1-w}\medskip
\end{minipage}
\begin{minipage}[b]{0.32\linewidth}
  \centering
  \centerline{\includegraphics[height=2.4cm]{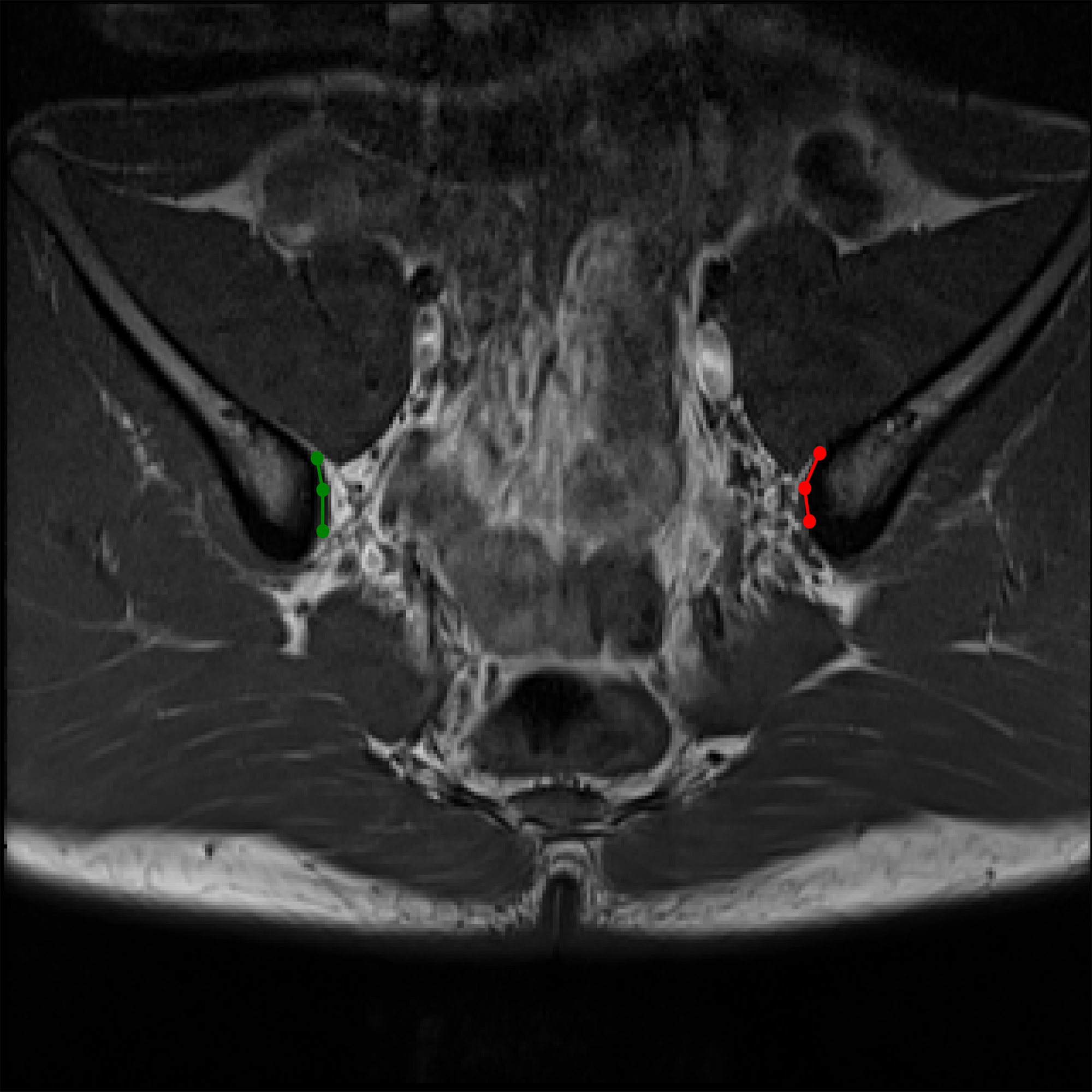}}
  \centerline{(e) Post. T1-w}\medskip
\end{minipage}

\begin{minipage}[b]{0.32\linewidth}
  \centering
  \centerline{\includegraphics[height=2.4cm]{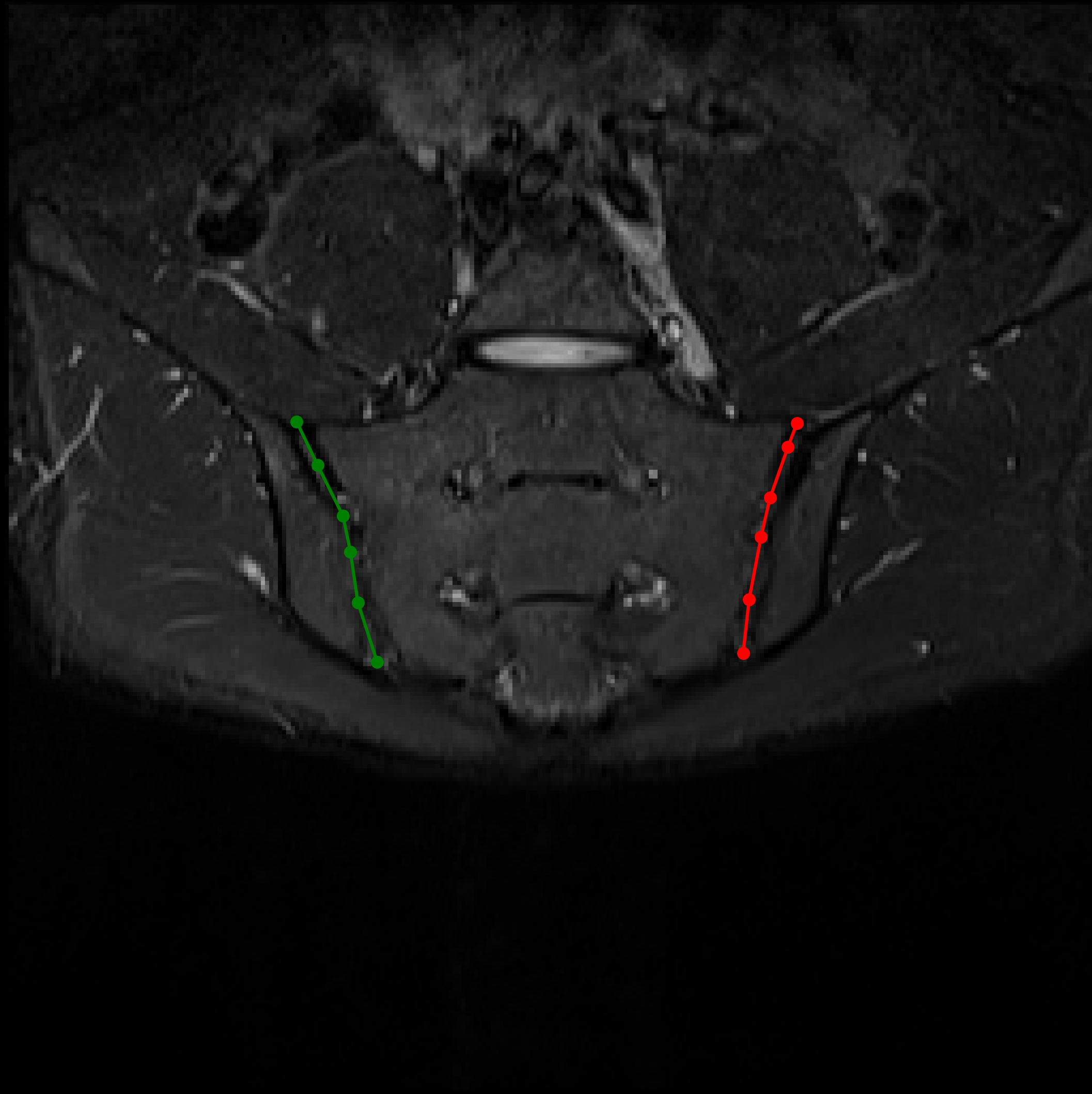}}
  \centerline{(e) Mid-cor. T2-w}\medskip
\end{minipage}
\begin{minipage}[b]{0.32\linewidth}
  \centering
  \centerline{\includegraphics[height=2.4cm]{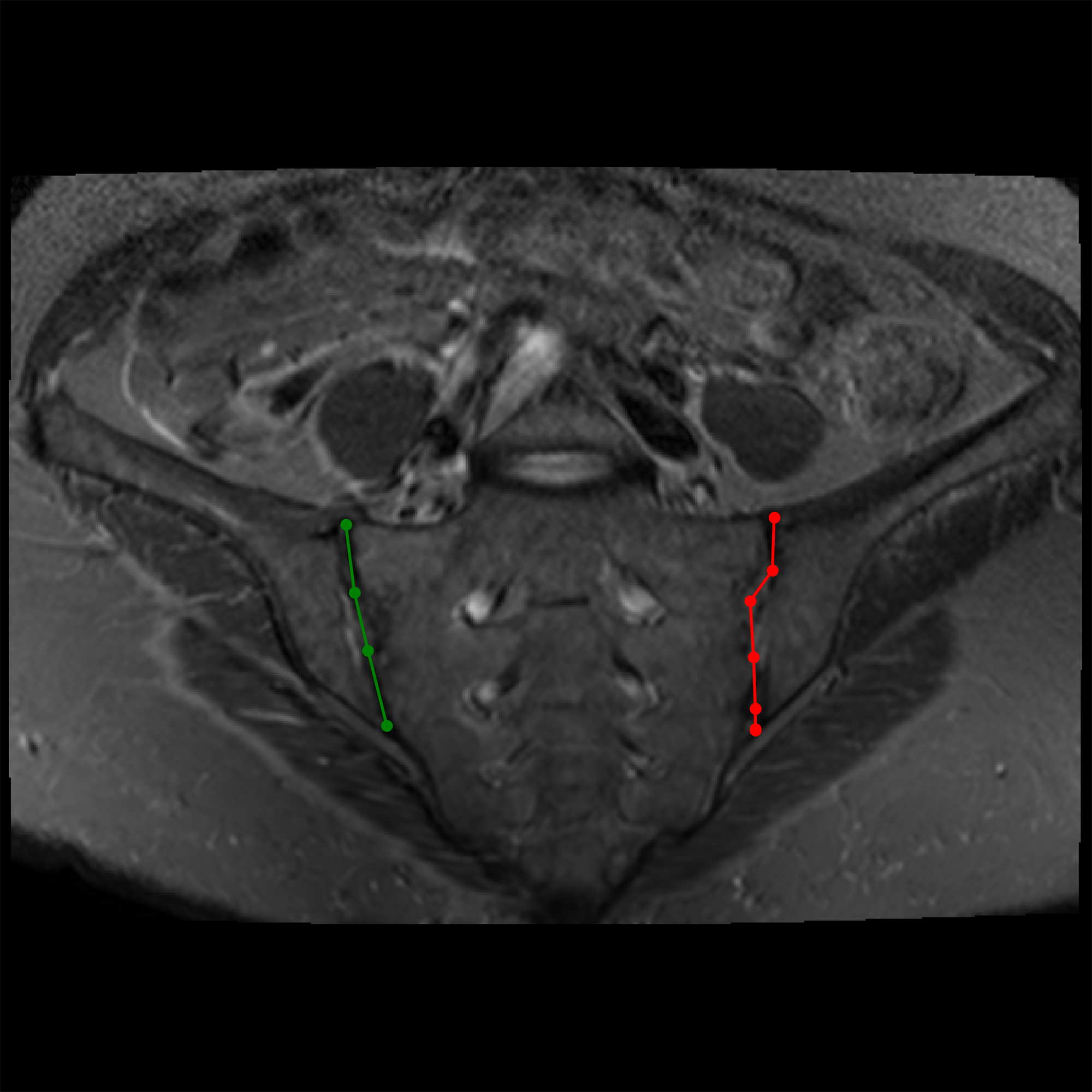}}
  \centerline{(e) Mid-cor. STIR}\medskip
\end{minipage}
\begin{minipage}[b]{0.32\linewidth}
  \centering
  \centerline{\includegraphics[height=2.4cm]{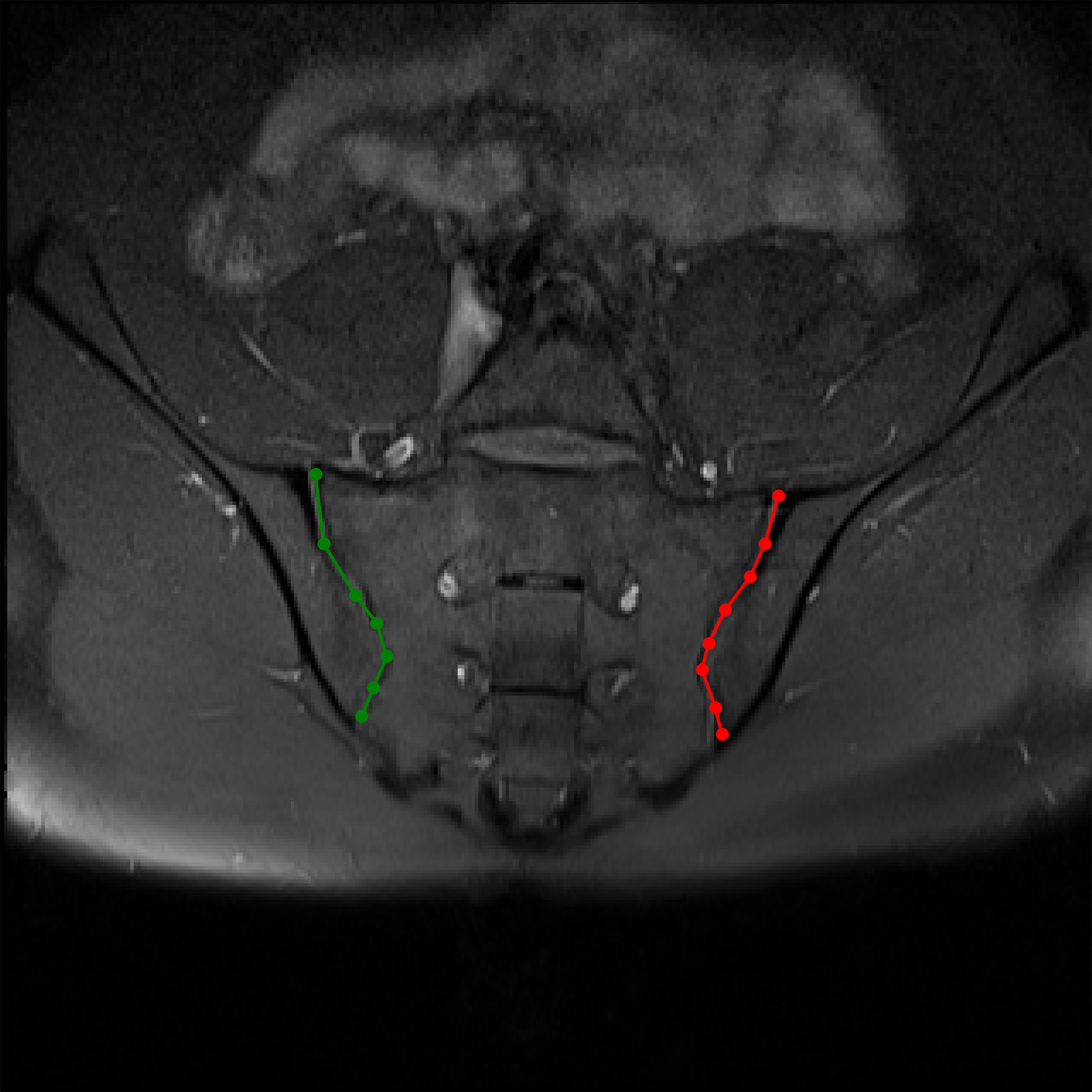}}
  \centerline{(f) Mid-cor. FS}\medskip
\end{minipage}





\caption{Example scans in the dataset with their marked-up annotated landmarks. (a), (b), and (c) are slices from the same T1-weighted scan at differing slice positions (anterior, mid-coronal, posterior) while (d), (e), and (f) are mid-coronal examples of different sequences in the dataset.}
\vspace{-1.25em}
\label{fig:data_anno}
\end{figure}

\vspace{-0.25em}
\section{Dataset \& Training Details}
\label{sec:dataset}

\noindent \textbf{Dataset.}
\label{ssec:dataset}
The Oxford Sacroiliac Joint (\textbf{OSIJ}) dataset is a collection of SIJ MRIs from 339 patients that have undergone scanning for in the Oxford University Hospitals NHS trust.
For experiments conducted in this work, the dataset is split into training (80\%), validation (10\%) and testing (10\%) sets on a per subject basis (271:34:34). Each subject possesses an average of two sequences (typically T1, T2, STIR, and FS) resulting in a total of 793 scans. Each scan roughly consists of 20 2D slices, resulting in a total of 16,978 images.

For the annotations of the contour of the SIJs, an expert was tasked with marking the landmarks (vertices) that best define both left and right SIJs through every slice in a given scan. The number of landmarks varies depending on the view of the SIJ; typically, mid-coronal SIJs cover a bigger image area demanding a larger number of landmarks and vice versa. The number of landmarks per slice ranges from 2 to 21.

\begin{figure}[!t]
  \centering
  \centerline{\includegraphics[width=\linewidth]{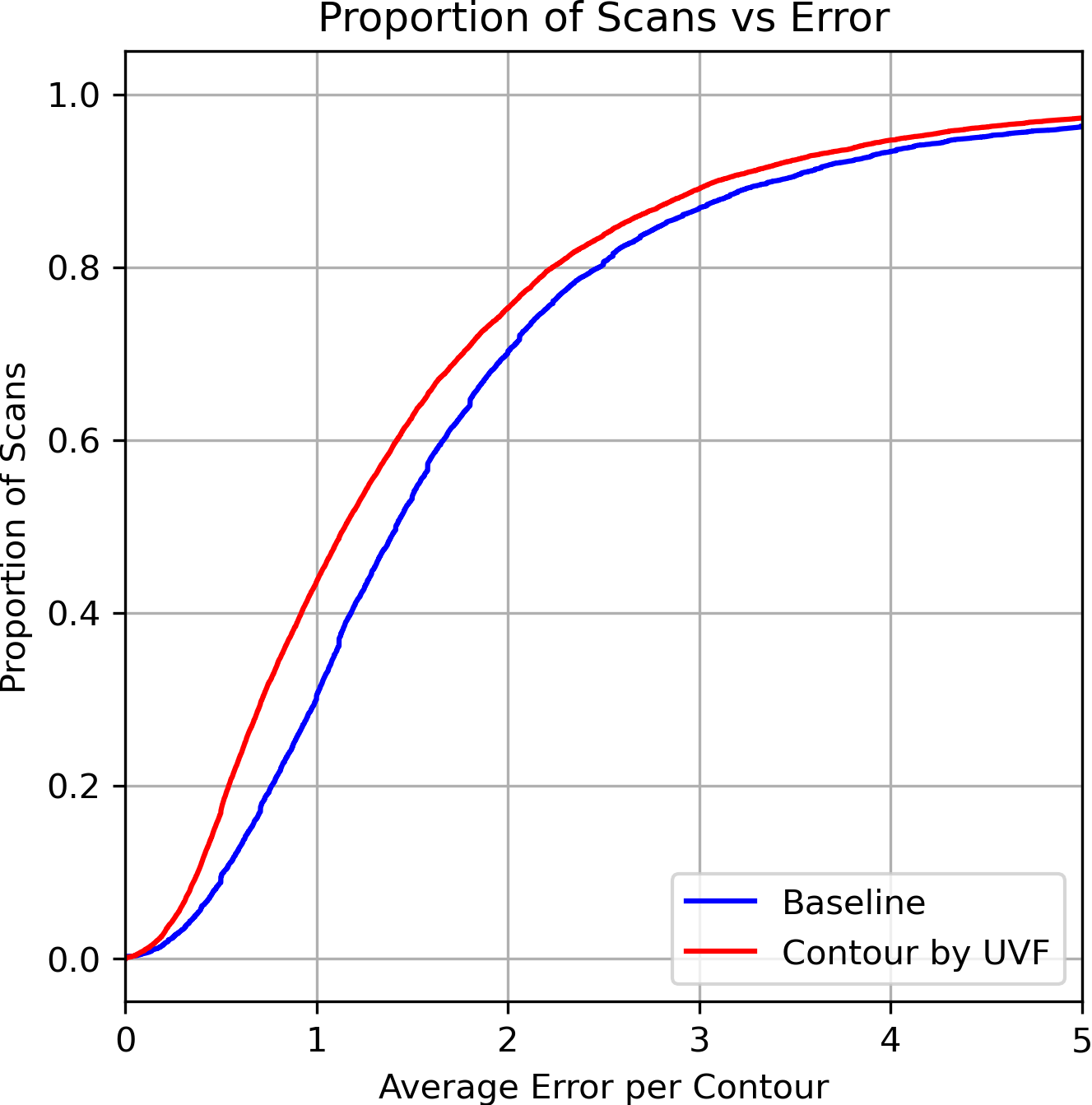}}
\caption{Cumulative test set error distribution (measured in pixels). Baseline is in blue and contouring via UVF is in red.}
\label{fig:errorprop}
\vspace{-1.0em}
\end{figure}

\hfill \break
\noindent \textbf{Training Details.}
\label{ssec:training}
The experiments in this work were conducted using a simple U-Net architecture~\cite{Ronneberger15}. For each contour, the network predicts 2 Gaussian heatmaps and 2 components (x and y direction) of the unit vector field; separate contours were predicted for each of the two SIJ (left and right). The SIJs are not guaranteed to be inside the field-of-view of the scans and as such these cases were kept in the training set to suppress false positives. The scans were typically squares in shape; thus, they are bi-cubically re-sampled to $224 \times 224$ pixels. Slices that were not square were padded with zeros prior to re-sampling so as to not change the aspect ratio.

The network is trained using an Adam optimiser \cite{kingma2014} with $\beta_1 = 0.9$, $\beta_2 = 0.999$ and a learning rate of $10^{-3}$ until convergence. Several augmentations were applied during training, namely: (a) translation $\pm20 \%$, (b) scale $\pm20 \%$, (c) rotation $\pm15 \degree$, (d) left/right flips, (e) additive Gaussian noise, and (f) Gaussian blur. A combination of L2-loss, for the UVF, and weighted L2-loss (see \cite{Windsor20}), for the Gaussian heatmaps, is used to train the network.

\begin{figure}[!t]
\begin{minipage}[b]{1.0\linewidth}
  \centering
  \centerline{\includegraphics[width=8.5cm]{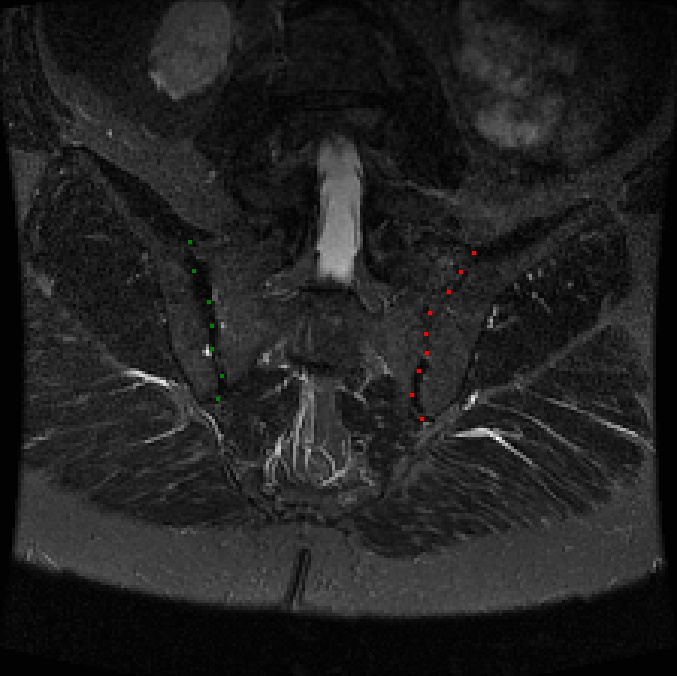}}
  \centerline{(a) Example with GT annotations}\medskip
\end{minipage}
\begin{minipage}[b]{.24\linewidth}
  \centering
  \centerline{\includegraphics[width=2.0cm]{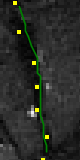}}
  \centerline{(b) Baseline}\medskip
\end{minipage}
\begin{minipage}[b]{0.24\linewidth}
  \centering
  \centerline{\includegraphics[width=2.0cm]{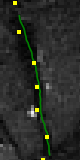}}
  \centerline{(c) UVF}\medskip
\end{minipage}
\begin{minipage}[b]{0.24\linewidth}
  \centering
  \centerline{\includegraphics[width=2.0cm]{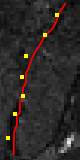}}
  \centerline{(d) Baseline}\medskip
\end{minipage}
\begin{minipage}[b]{0.24\linewidth}
  \centering
  \centerline{\includegraphics[width=2.0cm]{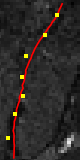}}
  \centerline{(e) UVF}\medskip
\end{minipage}

\caption{Quantitative result of the baseline against the proposed method on a test set example. Green contours highlight the right SIJ and red contours highlight the left; GT in yellow. (b) and (d) are from the baseline model while (c) and (e) are contours using UVF. Baseline predictions are sparse, with 21 landmarks for each contour, resulting in more aliasing.}
\vspace{-0.5em}
\label{fig:res}
\end{figure}

\begin{figure}[!t]
\begin{minipage}[b]{.48\linewidth}
  \centering
  \centerline{\includegraphics[height=3.6cm]{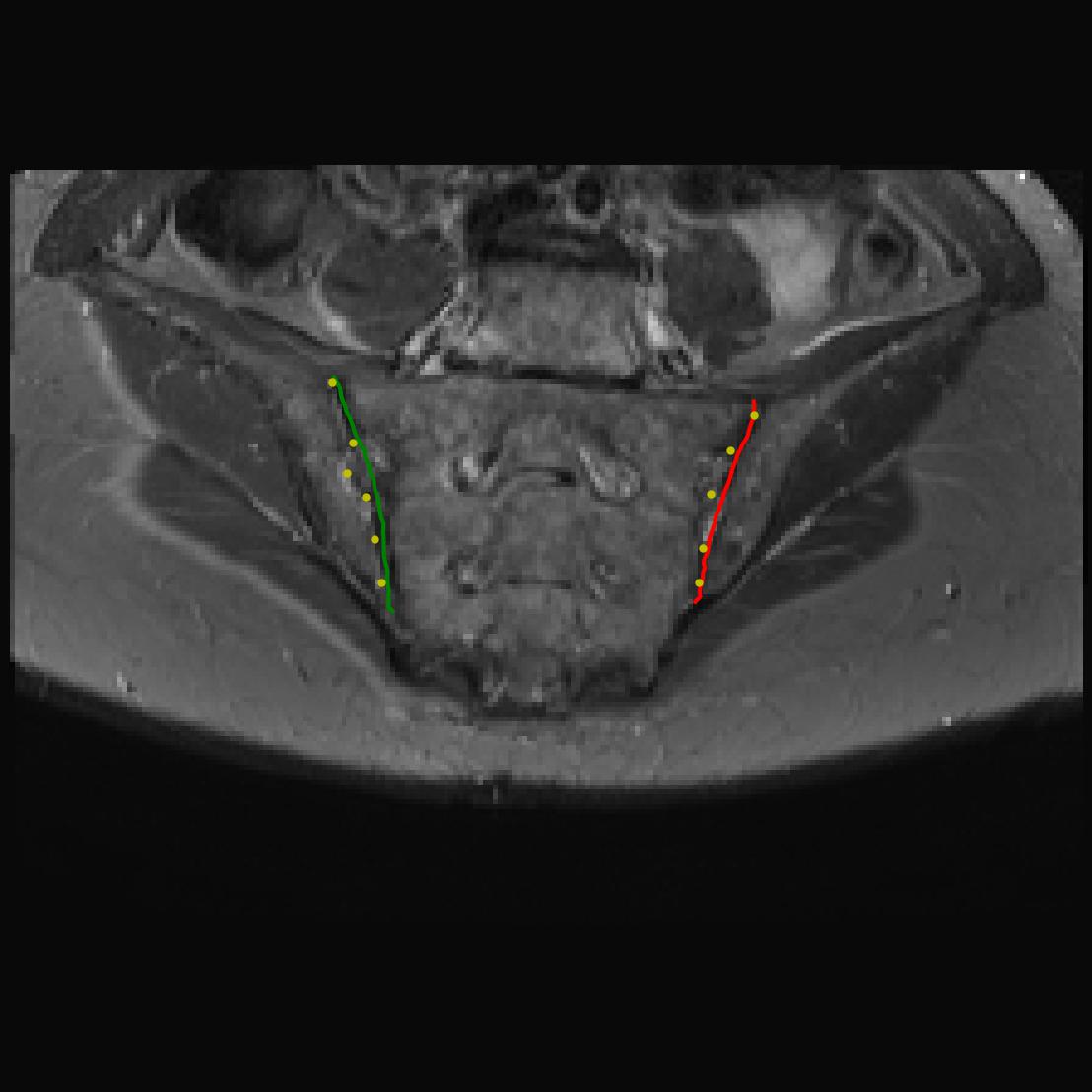}}
  \centerline{(a) \textbf{OSIJ} Test Sample 1}\medskip
\end{minipage}
 \hspace{0.2cm}
\begin{minipage}[b]{0.48\linewidth}
  \centering
  \centerline{\includegraphics[height=3.6cm]{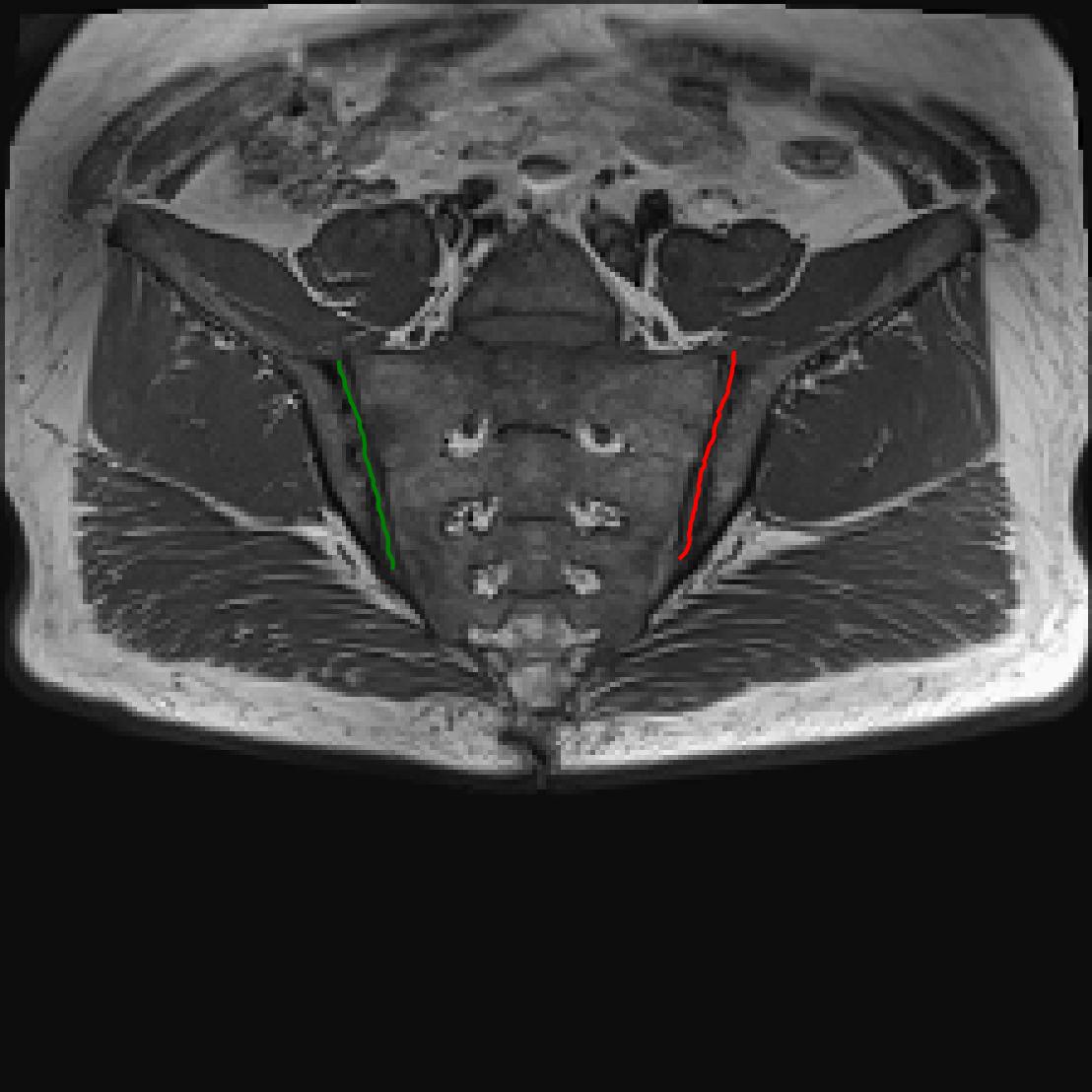}}
  \centerline{(b) rID: 73884}\medskip
\end{minipage}

\begin{minipage}[b]{0.48\linewidth}
  \centering
  \centerline{\includegraphics[height=3.6cm]{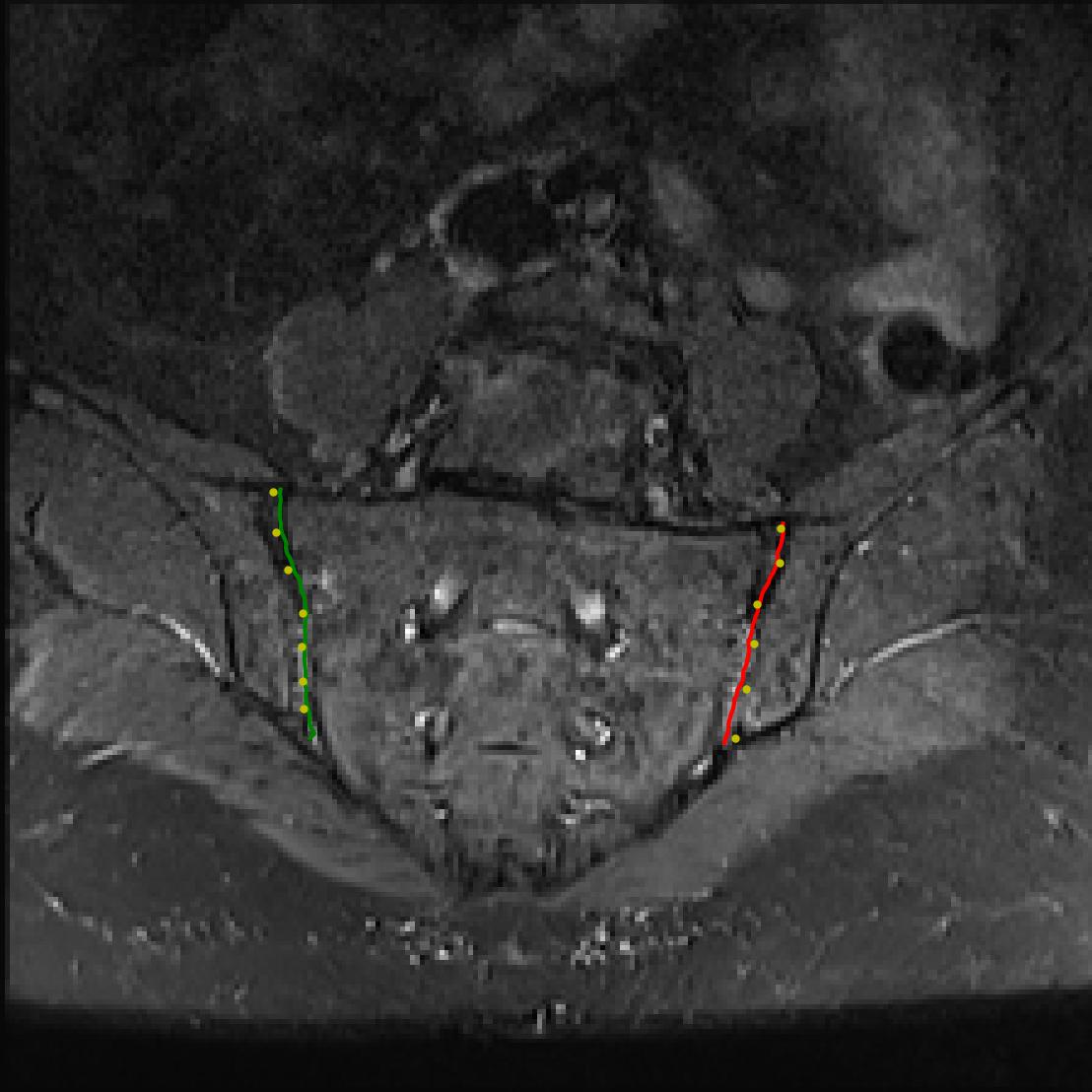}}
  \centerline{(c) \textbf{OSIJ} Test Sample 2}\medskip
\end{minipage}
 \hspace{0.2cm}
\begin{minipage}[b]{0.48\linewidth}
  \centering
  \centerline{\includegraphics[height=3.6cm]{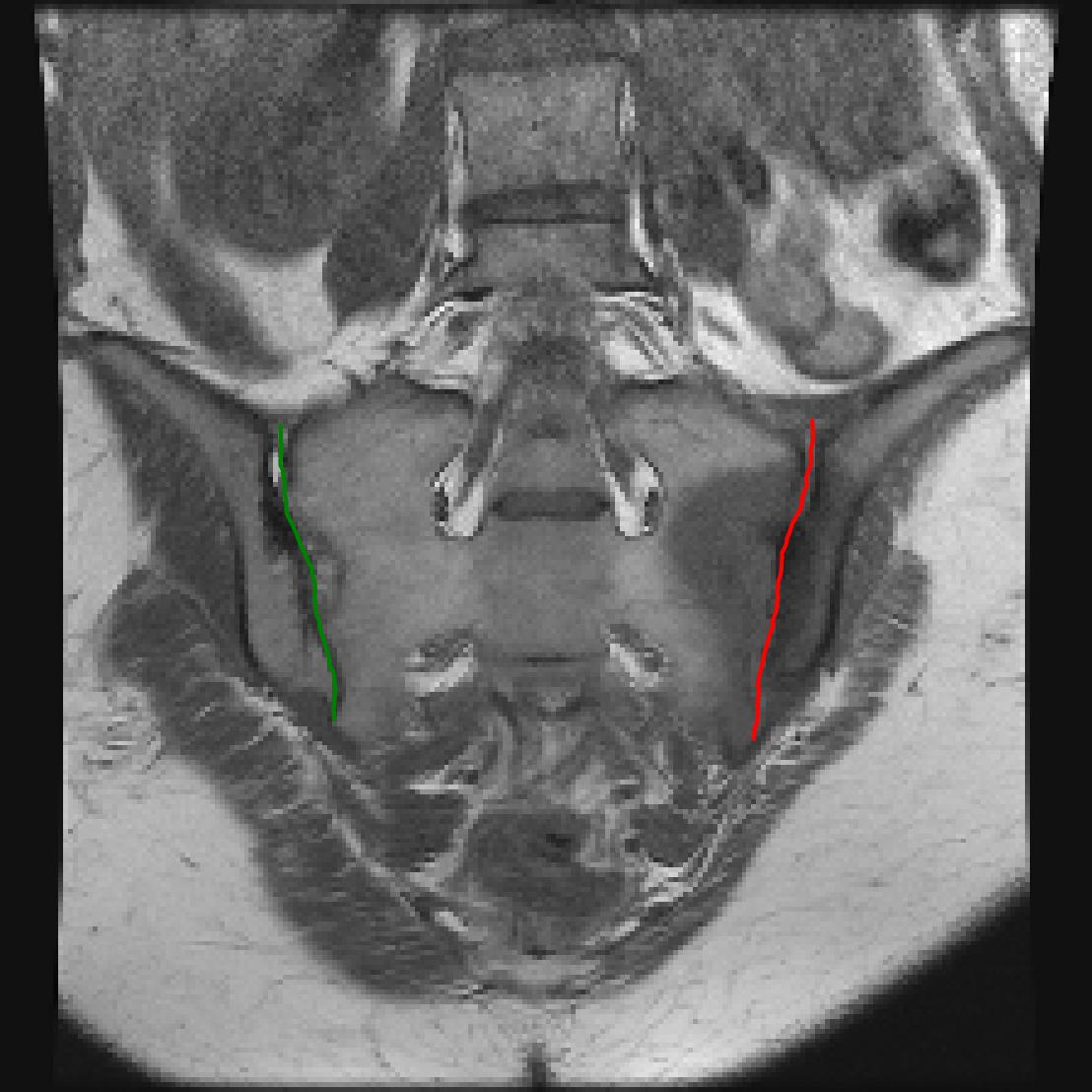}}
  \centerline{(d)  rID: 75292}\medskip
\end{minipage}

\begin{minipage}[b]{0.48\linewidth}
  \centering
  \centerline{\includegraphics[height=3.6cm]{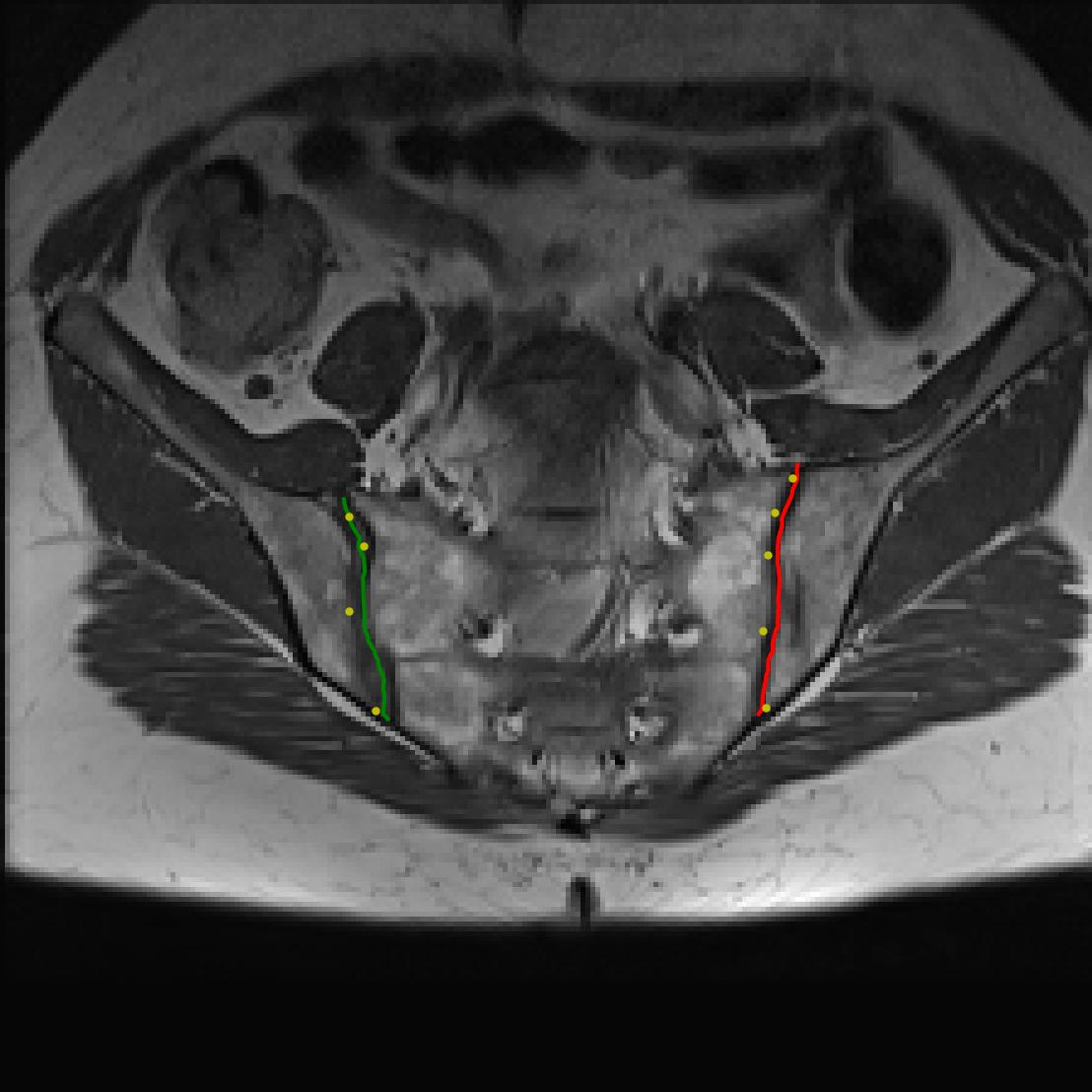}}
  \centerline{(e) \textbf{OSIJ} Test Sample 3}\medskip
\end{minipage}
 \hspace{0.2cm}
\begin{minipage}[b]{0.48\linewidth}
  \centering
  \centerline{\includegraphics[height=3.6cm]{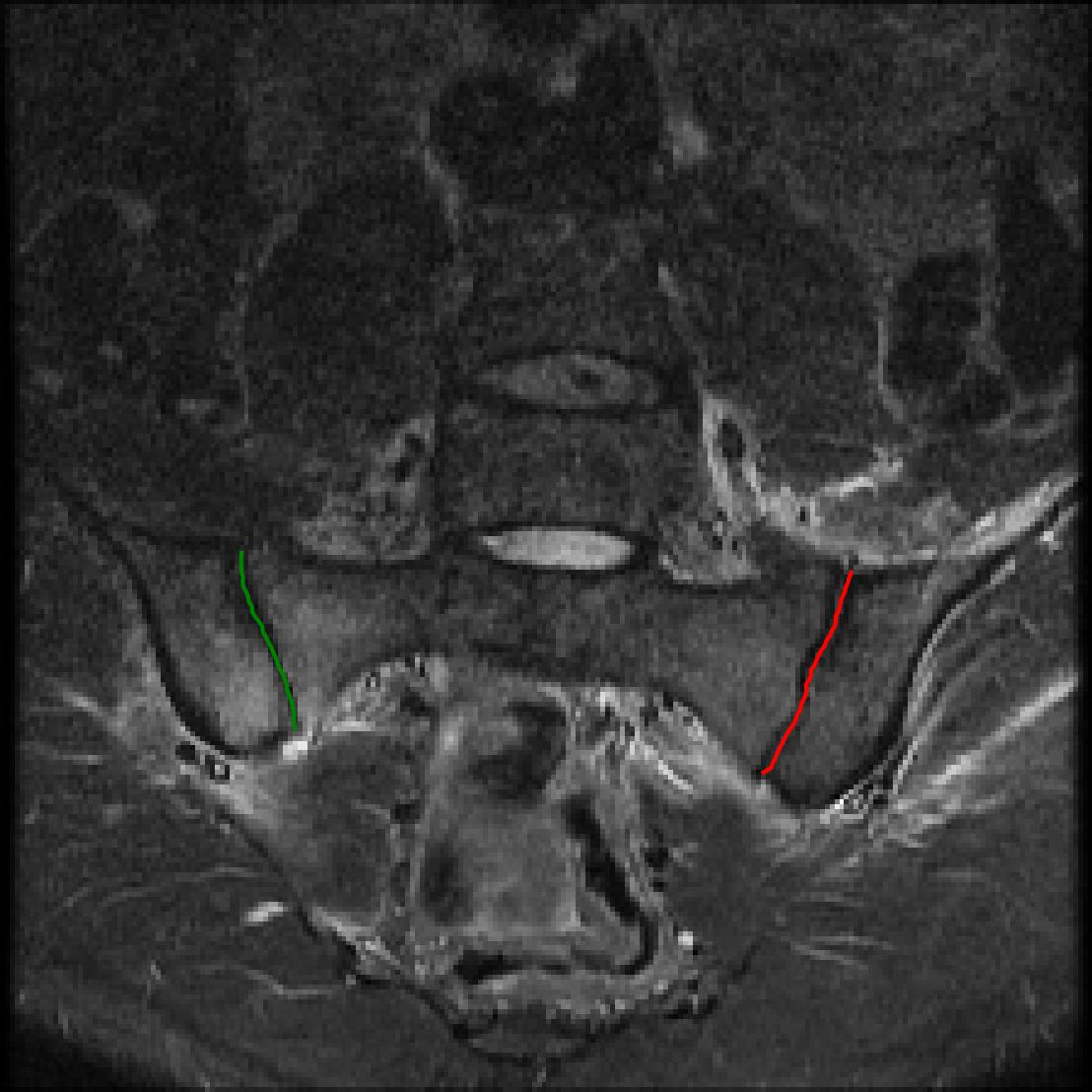}}
  \centerline{(f) rID: 154033}\medskip
\end{minipage}

\caption{Example contours via UVF. (a), (c), and (e) are from the \textbf{OSIJ} test with ground truth annotations in yellow (b), (d), and (e) are real-world unseen samples taken from Radiopaedia (73884, 75292, 154033).}
\label{fig:quant_res}
\end{figure}

\vspace{-0.25em}
\section{Performance Evaluation \& Results}
\label{sec:results}

For comparison we compare against a baseline network trained to predict 21 Gaussian heatmaps for each SIJ, 21 being the maximum number of landmarks in the dataset. We find this to be the simplest na\"ive solution to predict landmarks using a similar U-Net architecture as our proposed UVF approach. Samples with a lower number of annotated points were up-sampled via linear interpolation. At test time, each  prediction is compared against the ground truth landmarks of the contour and the root mean square (RMS) error is calculated from the closest points between the prediction and ground truth.

Results for both networks are shown in Figure~\ref{fig:errorprop} and Table.~\ref{tab:my-table}. Contouring by UVF overall works slightly better than the baseline ranging from 0.14 to 0.35 difference in RMS pixel error up to $95\%$ of the data in the test set. This might not seem like a huge amount, but Figure~\ref{fig:res} highlights that there is lower aliasing when looking at the contours using via UVF compared to just predicting landmarks via heatmaps. In general, $95\%$ of the test set have a lower than 4.5 pixel error which for our purposes is adequate for further downstream tasks e.g. defining an ROI for SIJ oedema classification. Figure~\ref{fig:quant_res} shows results on several examples both from \textbf{OSIJ}'s test set and to images extracted from Radiopaedia.

\begin{table}[!htb]
\begin{tabular}{ll|llllll}
\multicolumn{2}{l|}{Data Proportion} & 0.1 & 0.3 & 0.5 & 0.7 & 0.9 & 0.95 \\ \hline
\multicolumn{2}{l|}{Baseline Error}  & 0.52 & 1.00 & 1.41 & 2.00 & 3.40 & 4.45  \\
\multicolumn{2}{l|}{UVF Error}       & 0.38 & 0.72 & 1.15 & 1.76 & 3.10 & 4.10 
\end{tabular}
\caption{Table of RMS per proportion of data in the test set.}
\label{tab:my-table}
\end{table}

\vspace{-0.25cm}
\section{Conclusion}
\label{sec:con}
In this paper, we presented a pipeline to contour, focusing more on open contours but applicable to closed contours as well, objects in images and demonstrated its use to delineate SIJs in coronal spinal MRIs. Overall, the performance is better than the naive baseline of predicting landmarks of contours and is applicable to other contouring problems in medical image analysis.

\section{Compliance with Ethical Standards}
\label{sec:compliance}
The scans in the dataset were sourced from retrospective scans in the Oxford University Hospitals approved by the Health Research Authority (IRAS Project ID 207858).

\section{Acknowledgments}
\label{sec:acknowledgments}
The authors would like to thank Aimee Readie and Gregory Ligozio for their useful discussion and feedback during the research of this paper.
Rhydian Windsor is supported by CRUK as part of the EPSRC AIMS CDT (EP/L015897/1).

\bibliographystyle{IEEEbib}
\bibliography{strings,refs}

\end{document}